# Time-Efficient Hybrid Hyperparameter Tuning Approach for Cardiovascular Disease Classification

Abhay Kumar Pathak[1,#], Mrityunjay Chaubey[1,2,#,*], Manjari Gupta[1]

[1] *Department of Computer Science, Institute of Science, Banaras Hindu University, Varanasi, India.*
[2] *School of Computer Science, University of Petroleum and Energy Studies, Dehradun, India.*

[*] *Corresponding Author – mrityunjay.chaubey11@bhu.ac.in*

*ORCID: 0000-0001-5565-1933*
[#] *Equal contribution first author*

**Abstract**

Cardiovascular diseases (CVDs) are any serious illness of the heart, which require accurate diagnosis to prevent fatal consequences. Hyperparameter tuning plays a critical role in optimizing machine learning model performance by selecting the most suitable parameter configurations for improved accuracy, generalization, and reliability. Grid search systematically evaluates predefined hyperparameter combinations, whereas random search samples configurations randomly from the search space enabling broader exploration with reduced computational cost. Therefore, an efficient tuning strategy is essential when developing classification models where time plays an crucial role along with the predictive capability. In this work, we propose a new hyperparameter tuning approach to tune the hyperparameters of ML models for CVD classification. The proposed random grid search combines the power of random search to explore the global space with the focused and exhaustive search of grid search in the most promising areas. This hybrid approach finds an optimal balance between exploration and exploitation and yields a robust and time-efficient ML model for classification seetings. Experimental results on state of the art models demonstrated that randomised grid search performed better than traditional hyperparameter tuning methods. In addition to the observed improvement in model performance, the computational time required for training models was substantially reduced across most of the models. Presented results of the proposed study emphasizes the reduction in training time and computational efficiency of the proposed Randomized-Grid Search method. The proposed technique has significant potential to advance ML application in healthcare providing timely and accurate CVDs diagnosis.

## 1. Introduction

Cardiovascular diseases (CVDs) are the leading cause of mortality and morbidity globally, including approximately 17.9 million lives every year, which cumulates to 31% of global deaths [1]. The consistent increase in CVDs prevalence, influenced by factors such as aging, irregular blood pressure, and the population's lifestyle changes, along with the regional factors, also has a significant role in it. Comorbidities like diabetes and hypertension have led to an increasing burden on healthcare systems worldwide, which necessitates early and accurate diagnosis to mitigate the impact of heart disease by allowing timely intervention to reduce the mortality rate and economic burden of healthcare [2]. However, there are traditional diagnostic methods for heart diseases; they rely heavily on manual interpretation of electronic health records. Subtle trends and interactions that are clear evidence of progression of heart disease can be missed [3-5]. CVDs are very serious and this challenge needs to be overcome by automating the process of diagnosis. Therefore, it generated the interest in the application of artificial intelligence (AI) that led to the development of more accurate and automated diagnostic tools for CVDs [6-8]. Machine Learning (ML) has demonstrated promising results in improving the performance of heart disease classification. ML models can analyse large sets of electronic health records to detect complex patterns that can be imperceptible to clinicians [9-16]. State of the art models like Random Forest (RF), K-Nearest Neighbour (KNN), Decision Tree (DT), Extreme Gradient Boosting (XGB) and Multilayer Perceptrons (MLPs) have demonstrated their ability in disease risk prediction, early diagnostics and classification of longitudinal health data [17]. Tree based models have been popular among different ML classification models as they can handle large dataset, are robust in maintaining balance on both categorical and numerical attributes and are resistant to overfitting [18-23].

The ML models are generally usable but are very sensitive to the correct choice of hyperparameters. The performance of ML models is very reliable and depends on the selections of appropriate hyperparameters values. Hyper-parameters of the ML models have a great impact on the performance, generalisability and computational efficiency of the models [24-25]. Hence, tuning these hyperparameters is required to get the best performance. However, finding the best set of parameters is often difficult and computationally expensive in terms of time and space complexity, particularly for a large number of mutually dependent parameters.

Hyperparameter tuning of ML models is generally performed using random and grid search [26]. Grid search involves an exhaustive search over a predefined set of hyperparameters, attempting all possible combinations to find the best hyperparameters for optimal model performance [27]. This approach is effective but can be computationally expensive especially when the requirement of hyperparameter space is too large. On the other hand, Random Search makes use of the random combinations of hyper-parameters from a pre-defined space which is shifted to the exploitations over the separation leading to it. This approach improves efficiency but often ignores the best hyperparameter set combinations [28]. In contrast, it selects random combinations to omit the optimal set of hyperparameters to promise the efficiency but does not guarantee the comprehensive exploration of the entire search space. Recent research has introduced techniques that integrate random sampling with some targeted searches in promising regions of the hyperparameter space, allowing more efficient optimization of the ML models [29-34]. The trade-off between exploration and exploitation led us to develop a hybrid approach that combines the strength of exploration and the efficiency of random search together.

In this study, a novel hybrid search has been proposed called Randomized-Grid Search to address the limitation of traditional random and grid searches for hyperparameter tuning to yield the best probable results. It efficiently searches the most promising areas, while the random search provides an exhaustive search in the targeted regions. With an extensive list of available hyperparameters that can be fine-tuned to avoid overfitting and enhance generalizability. Our experiment was conducted on the 'UCI Heart Disease' dataset, which demonstrated that the randomized-grid search algorithm outperforms both traditional random and grid searches in terms of different performance evaluation matrices. Additionally, it significantly reduces the computational cost of hyperparameter tuning while providing better performance. This makes Randomized-Grid search a valuable tool for optimizing ML models in real-world applications where time and computational resources are often limited and the performance of the model is crucial. Our major contributions are as follows:

i. Proposed Randomized-Grid Search approach for hyperparameter tuning that utilizes random exploration and localized grid refining for effective optimization.

ii. We validate the same tuning procedure across classical ML and ensemble classifiers to show model-agnostic applicability.

iii. Several State-Of-The-Models on UCI Heart Disease dataset exhibit consistent accuracy and ROC-AUC gains over Grid Search and Random Search.

iv. Shows high computational efficiency for expensive models reducing tuning time relative to Grid Search while enhancing performance.

## 2. Related Work

Machine learning (ML) has become very popular in the field of disease prediction. It has shown promising results in diagnosing a wide range of conditions, from Parkinson's disease to cancer and cardiovascular diseases. Hyperparameter tuning is a very important step in the machine learning process. It directly affects how well a model works by changing things like learning rates, regularization factors, and settings that are specific to the model.

| **Recent Key Contributions** | **Methodology** | **Proposed Methodology** | **Advantages** | **Limitations** |
| --- | --- | --- | --- | --- |
| Parkinson's Diagnosis with Generalized Artificial Bee Colony [35] | ResNet50 for Parkinson’s disease diagnosis optimized by Generalized Artificial Bee Colony (ABC) algorithm | Integration of Artificial Bee Colony with Range pruning and Dimension adjustment strategies for ResNet50 hyperparameter tuning | High accuracy (greater than 96%) on the Mobile Device Voice Recordings at King’s College London dataset, computationally efficient | Limited to speech-based features, may not generalize to other data modalities |
| Respiratory Disease Prediction using Hyperparameter Optimization [36] | Hybrid model with hyperparameter optimization using Genetic Algorithm and binary Grey Wolf Optimization for feature selection | Genetic algorithm-based hyperparameter optimization with an ensemble model | Improved accuracy in disease prediction, enhanced efficiency in hyperparameter tuning | High computational cost in feature selection phase, limited to respiratory diseases |
| Cardiovascular Disease Prediction with Two-Phase Taguchi Method, Hyperparameter Artificial Neural Network, and Genetic Algorithm [37] | Two-phase Taguchi Method combined with Genetic Algorithm and Artificial Neural Network for hyperparameter optimization | Optimizing hyperparameters using the Two-Phase Taguchi Method, Hyperparameter Artificial Neural Network, and Genetic Algorithm framework for Artificial Neural Network models | Requires fewer experiments, efficient for low-power devices, good performance with limited resources | Relies on Artificial Neural Network models, which might not generalize well across different Cardiovascular Disease datasets |
| Agricultural Disease Recognition with Lightweight Adaptive Velocity Particle Swarm Optimization [38] | RepAgrViT, a hybrid Convolutional Neural Network and Vision Transformer architecture with Lightweight Adaptive Velocity Particle Swarm Optimization for hyperparameter optimization | A lightweight Vision Transformer model with efficient parameter fusion and neighbourhood-based local search strategies | Achieves high accuracy (98.54%) while maintaining computational efficiency | Potential overfitting in small datasets, complex hybrid models require detailed tuning |
| Lung and Colon Cancer Classification using Deep Learning Ensemble Approach with Explainable Artificial Intelligence [39] | Deep learning ensemble model using Xception and MobileNet for cancer diagnosis with Gradient-weighted Class Activation Mapping for interpretability | Combines Xception and MobileNet in an ensemble approach with visual explanations via Gradient-weighted Class Activation Mapping | High classification accuracy (99.44%), enhanced model interpretability for clinical trust | Limited generalizability across different cancer datasets, computationally expensive |

***Table 1:*** *Recent key literatures with their advantages and limitations*

In this section, we examine several cutting-edge techniques for hyperparameter tuning used in disease prediction as shown in Table (1), emphasizing the advantages and drawbacks of current methods. Wang et al. (2023) worked on the diagnosis of Parkinson's disease using the ResNet50 model optimised by a Generalised Artificial Bee Colony (ABC) algorithm [35]. This approach used state-of-the-art techniques like Range Pruning and Dimension Adjustment to enhance the hyperparameter tuning. The method was very efficient in using computers as it achieved more than 96% accuracy on the Mobile Device Voice Recordings at King's College London dataset. However, it could only use speech-based features, making it less useful for other types of diagnostic data, like medical imaging.

Kaur et al. (2024) proposed a hybrid model for prediction of respiratory disease [36]. It uses Genetic Algorithm (GA) for hyperparameters tuning and binary Grey Wolf Optimisation for feature selection. The method is more accurate and efficient and can be applied in resource-poor settings. However, the method was very expensive to compute in the feature selection phase and it can only be applied for respiratory diseases, which limits its use for other fields. Lin et al. (2024) proposed the Two-Phase Taguchi Method (TPTM) in combination with Hyperparameter Artificial Neural Network (HANN) and Genetic Algorithm (GA) for the optimisation of hyperparameters in prediction of cardiovascular diseases [37]. Their method needed less tests than the regular grid search and worked well for low power devices. But it heavily depends on artificial neural networks (ANNs) which may not perform well for different datasets. This makes it less useful in other forms of cardiovascular disease.

Dai et al. (2025) proposed RepAgrViT , a hybrid model consisting of a Convolutional Neural Network (CNN) and a Vision Transformer (ViT) [38]. It was refined with Lightweight Adaptive Velocity Particle Swarm Optimisation (Lite-AVPSO) to support IoT edge conditions to identify agricultural diseases. The model was very accurate (98.54%) and still worked fast, which is important for real time agricultural monitoring in places where resources are limited. But with small data set, the model may not work well as it may overfit them. Moreover, the complex hybrid model needs to carefully tune its hyper-parameters, which can be computationally expensive.

Vanitha et al. (2024) suggested a deep learning ensemble model for lung and colon cancer classification using Xception and MobileNet architectures, incorporating Gradient-weighted Class Activation Mapping (Grad-CAM) for model interpretability [39]. The ensemble achieved a good classification accuracy of 99.44%. The application of Grad-CAM guarantees the transparency of the diagnostic reasoning of the model, which is critical for acceptance in clinical practice. But this method is expensive to implement on a computer and may not work well on all cancer data set with different imaging conditions. The above-mentioned related works depict the importance of hyperparameter tuning for improving the prediction of diseases by machine learning models . However, these techniques often suffer from the drawbacks of high operational costs, limited applicability to specific models, and restricted use in diverse scenarios.

Grid Search and Random Search are both simple and widely used, but they don't do a good job of balancing exploration and exploitation, or they cost a lot of computing power when working with complex datasets. Bayesian optimisation and AutoML are more sophisticated methods requiring more computing power and hence are more difficult to deploy in resource-limited environments. Although disease prediction models have advanced, they still have major problems in terms of speed, usefulness for different types of diseases, and model complexity.

Our proposed Randomized-Grid Search technique is different from these other methods because it doesn't depend on the model. Instead, it uses a balanced approach that quickly searches hyperparameter spaces and then takes advantage of the best areas. Randomized-Grid Search is faster than traditional grid search, which is time-consuming and thorough, and it doesn't lose accuracy. This hybrid strategy can be used with many different types of machine learning models, such as decision trees, support vector machines, and deep learning models. It is a useful and effective way to predict diseases in healthcare and other fields. Also, Randomized-Grid Search fills the gap left by earlier studies by offering a scalable, resource-efficient method that guarantees better performance even in real-world settings where resources are limited.

## 3. Proposed Methodology

In this section, a hybrid approach (Randomized-Grid Search) has been presented for hyperparameter tuning which combines the strength of random search for exploration and grid search for exploitation. The proposed technique further applied and validated for CVDs classification problem which uses the UCI Heart Disease Dataset.

**Random Search:** A hyperparameter optimization technique which involves randomly selecting values of hyperparameters from the pre-defined space as shown in Fig. 1. Instead of evaluating all possible combinations like Grid Search, it selects sample points randomly.

Let Θ represent the hyperparameter space where each parameter $\theta_i$ drawn from the probability distribution $\mathcal{P}_j$:

$$\theta_i \sim \mathcal{P}_j$$

Given $n$ random samples from the hyperparameter space the goal is to find the optimal hyperparameter space $\theta^*$, that minimizes the specific loss function $\mathcal{L}$ . The mathematical expression of random search is given as:

$$\theta^* = arg_{\theta \in \{\theta_1, \theta_2, \ldots\ldots, \theta_n\}} \{\min \mathcal{L}(f_\theta(x_{train}), y_{train})\}$$

Where $f_\theta$ is the model parametrized by $\theta$, $x_{train}$ is the training data and $y_{train}$ is the corresponding label.

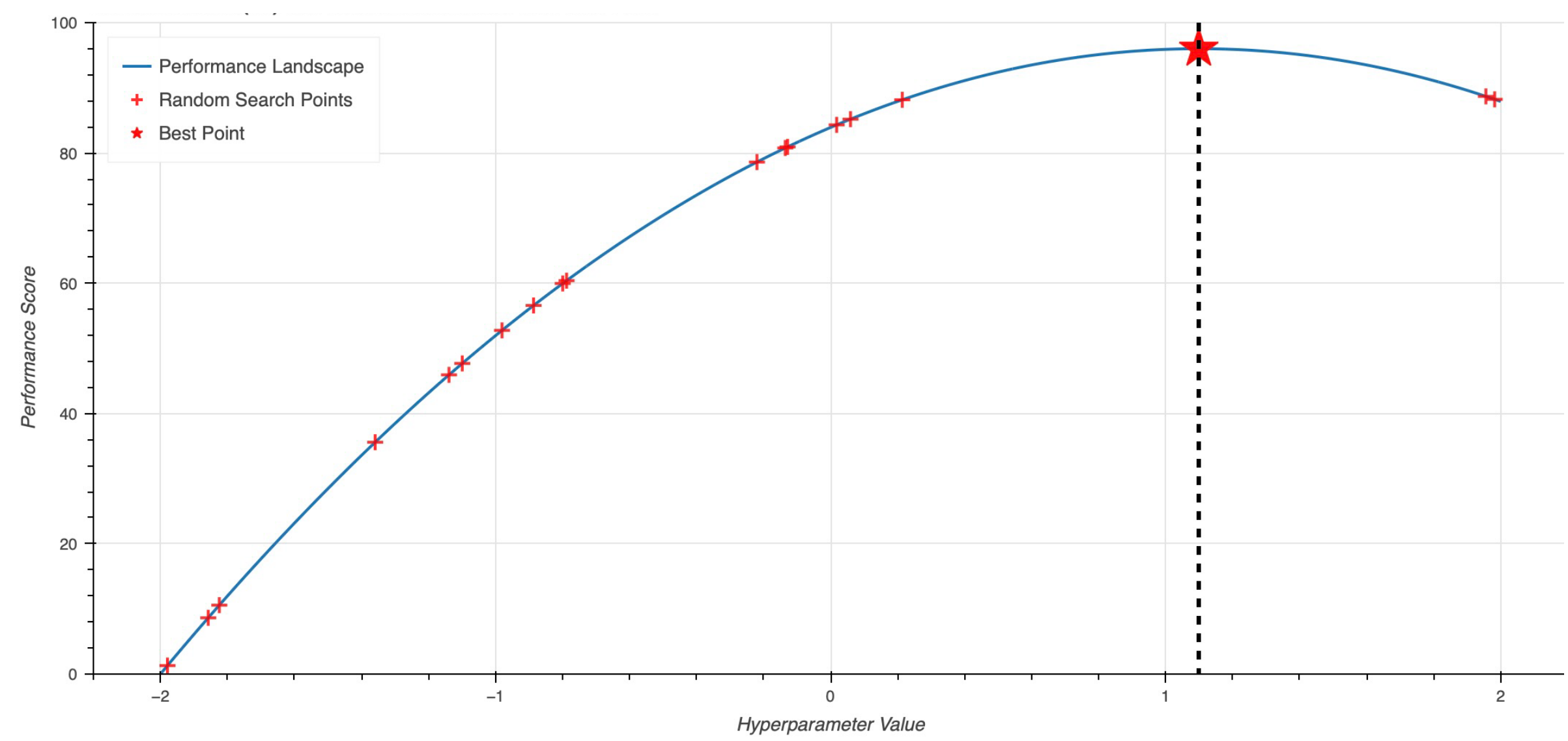

***Fig. 1:*** *Image illustrating the randomly selected search points in random search on a single hyperparameter. (a) Curve depicts the performance of a single hyperparameter values; (b) The dashed straight lines depicts the best performance point for hyperparameter; (c) Cross depicts the points taken by random search.*

**Grid Search**: A hyperparameter optimization technique which systematically explores the hyperparameter space by evaluating all possible combinations using predefined hyperparameter values as shown in Fig. 2.

Let $\Theta$ be a hyperparameter space of a discrete grid where each hyperparameter $\theta_i$ takes a specific value from the predefined set $\Theta_i$, It can represent in the form of cartesian product of the set of values of each parameter:

$$\Theta_{grid} = \theta_1 \times \theta_2 \times \ldots \times \theta_p,$$

Where $\theta_1, \theta_2 \ldots \theta_p$ represents the possible values for each of the p parameters. The optimization problem can be expressed as:

$$\theta^* = arg_{\theta \in \Theta_{grid}} \{ \min \mathcal{L}(f_\theta(x_{train}), y_{tarin}) \}$$

Where $\mathcal{L}$ is a loss function, $f_\theta$is the model parametrized by $\theta$, and $\Theta_{grid}$ represents all the possible combinations of hyperparameters in the grid.

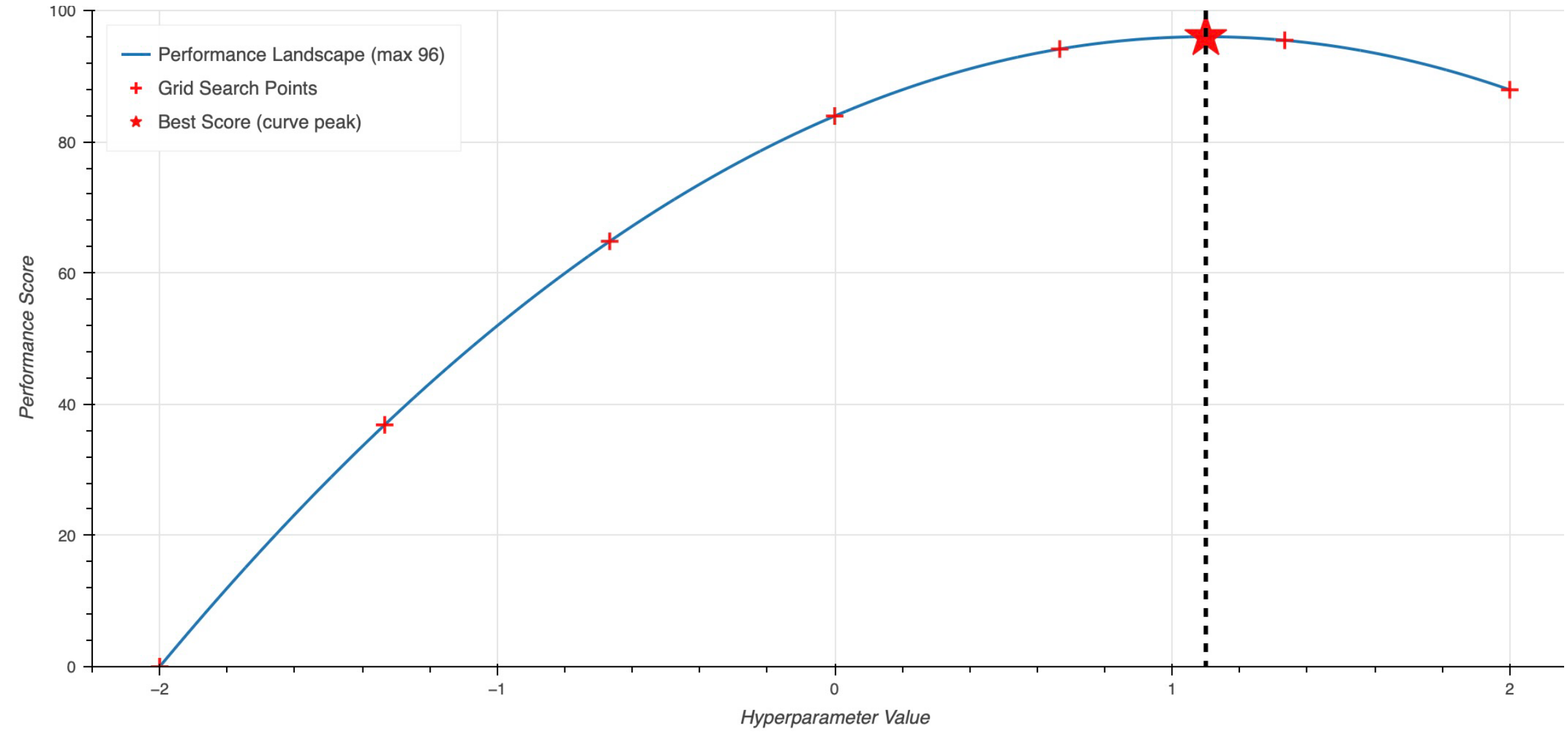

*__Fig. 2:__ Image illustrating the selected search points in grid search on a single hyperparameter. (a) Curve depicts the performance of a single hyperparameter values; (b) The dashed straight lines depicts the best performance point for hyperparameter; (c) Cross depicts the points taken by grid search.*

### 3.1. Mathematical Expression of proposed Randomized-Grid Search

Initially, randomly select $m$ sample of hyperparameter sets from the space $\Theta$ :

$$\hat{\theta}_{random} = arg_{\theta\in\{\theta_1,\theta_2,\dots,\theta_m\}}\{\min \mathcal{L}(f_\theta(x_{train}), y_{tarin})$$

Where $\theta_i \sim \mathcal{P}_j$ , where $\mathcal{P}_j$ is probability distribution of hyperparameter and $\hat{\theta}_{random}$ is best performing configuration of parameter values from random search.

After identifying $\hat{\theta}_{random}$, restrict the search space to smaller grid $\Theta_{grid}$, which is centred around $\hat{\theta}_{random}$, then exhaustive grid search is applied for fine tuning:

$$\theta^{*} = arg_{\theta\in\Theta_{grid}(\hat{\theta}_{random})}\{\min \mathcal{L}(f_\theta(x_{train}), y_{tarin})\}$$

Where $\Theta_{grid}(\hat{\theta}_{random})$ is narrowed search space which is presented by $[\hat{\theta}_{random} - \Delta_j , \hat{\theta}_{random} + \Delta_j\,]$ around $\hat{\theta}_{random}$ chosen by smaller variations $\Delta_j$ around the hyperparameter of $\hat{\theta}_{random}$.

The graph illustrates in Fig. 3, the process of Randomized-Grid hyperparameter tuning process which uses a combination of random and grid search. The X-axis shows the single hyperparameter values to illustrate the basic approach of the proposed technique and Y-axis shows the corresponding performance score. The model's performance all across the hyperparameter values are depicted by the green line. The randomly selected points are shown by the red cross, which demonstrates the randomly selected hyperparameter values over a broad range. After this step, random search identifies the high performing regions of hyperparameter values and grid is applied within yellow shaded area focusing on systematically exploring a smaller range of hyperparameter for further optimization.

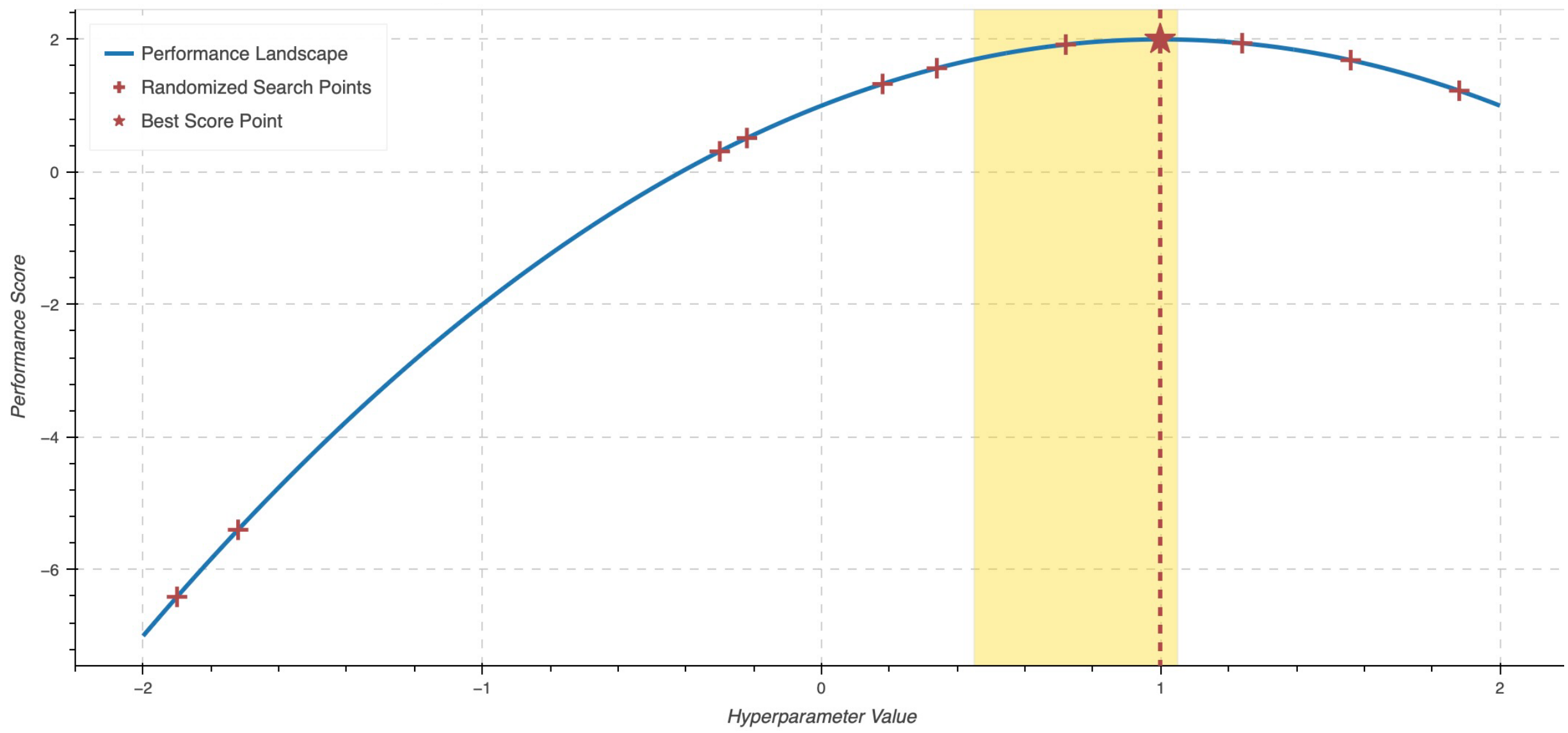

*__Fig. 3:__ Image illustrating the randomly selected search points with grid search on a single hyperparameter; (a) Curve depicts the performance of a single hyperparameter values; (b) Cross depicts the points taken by random search; (c) Subsequently the yellow window depicts the narrow range (neighbourhood values) of hyperparameter*

*values chosen by grid search; (d) Dashed straight line depicts the best performance point for the hyperparameter value.*

### 3.2. Integration of Randomized-Grid Search with State-Of-The-Art Models

The proposed framework presented in Fig. 4, illustrates the stepwise process of the model. It focuses on the pipeline used for CVDs classification using UCI Heart Disease dataset. The initial steps are Data Acquisition which have 13 independent and 1 dependent variable feature. Data Preprocessing involves Normalization, Null value imputation and outlier remover to prepare dataset for the model.

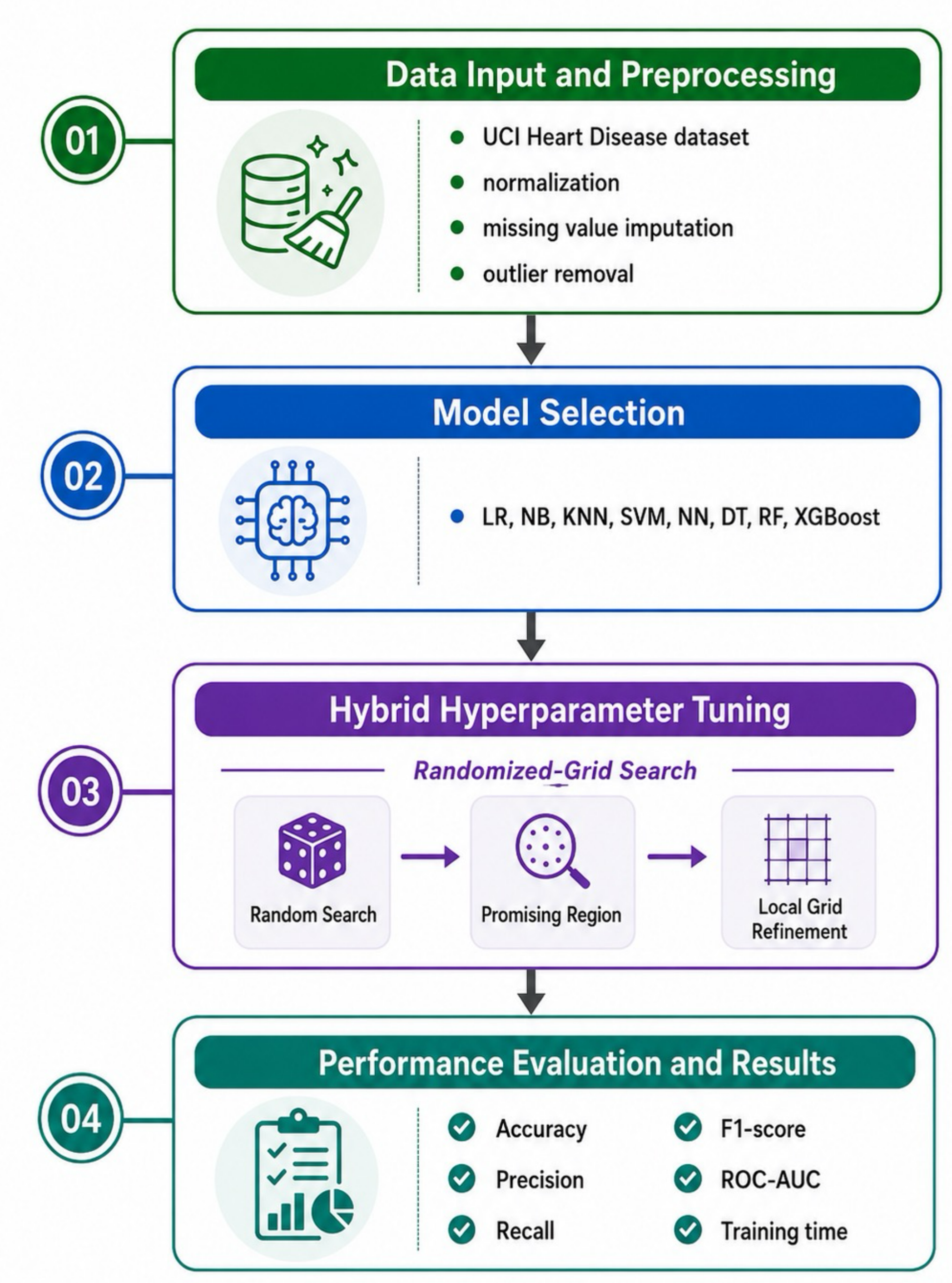

***Fig. 4:*** *Detailed pipeline of the proposed method.*

In next step, state-of-the-art machine learning models are employed with its selected hyperparameters. Following the ML classifier models, hybrid hyperparameter tuning has been employed which combines the strength of random search and Grid Search. In the next step, optimized model with best hyperparameter set of values assessed using performance evaluation metrices. In the last step, the results are shown with a focus on improved model generalizability and reduced computational cost.

Algorithm for Randomized-Grid search integration with selected machine learning model training, it aims to optimize the hyperparameters $\gamma_1, \gamma_2, \dots, \gamma_n$.

**Problem Setup**: Given the dataset $\mathcal{D} = \{(x_i, y_i)\}_{i=1}^{n}$where:

$x_i \in \mathbb{R}^d$represents the feature vectors with d dimensions,

$y_i \in \{0,1\}$ is binary class label for each instance $i$.

We defined hyperparameter space $\Gamma = (\gamma_1, \gamma_2, \dots, \gamma_n)$, where $\gamma_1, \gamma_2, \dots, \gamma_n$ are the ML classifier hyperparameters. The objective is to minimize the validation loss $\mathcal{L}$ by optimizing these parameters $\gamma_1, \gamma_2, \dots, \gamma_n$ through Randomized-Grid search.

**Step 1. Random Search**:

**1.1: Random Sampling**: Randomly selects $m$ samples of hyperparameters configuration $\Gamma = (\gamma_1, \gamma_2, \dots \gamma_n)$ from predefined probability distributions $\mathcal{P}_j$ of hyperparameter:

$$\gamma_n \sim \mathcal{P}_j, \qquad \forall\, j \in (1,2, \dots, n)$$

Where $\mathcal{P}_j$ is a probability distribution for each parameter. The distribution can be defined for each parameter according to the nature of hyperparameters.

**1.2: Model Fitting**: For each sampled hyperparameter configuration $\gamma_i$, train a ML classifier $f_{\gamma_i}$ on training dataset $\mathcal{D}_{Training}$:

$$f_{\gamma_i} = Train_ML_Classifier(\mathcal{D}_{Training}, \gamma_i)$$

Where $f_{\gamma_i}$ represents the ML classifier trained with $i^{th}$ hyperparameter from sample hyperparameters. The process of training involves creating a classification model using hyperparameter configurations $\gamma_i$.

**1.3: Model validation**: Compute the validation loss $\mathcal{L}^{(i)}$ for each model $f_{\gamma_i}$ on validation set $\mathcal{D}_{Validation}$:

$$\mathcal{L}^{(i)} = \mathcal{L}(f_{\gamma_i}\,(x_{Validation}), y_{Validation})$$

Where $\mathcal{L}$ represents the loss function. This step allows us to assess the performance of the model on unseen data for each configuration of $\gamma_i$.

**1.4: Storing Results**: Store each hyperparameter configuration and its corresponding loss:

$$\mathcal{H}_{Random} = \{\gamma_i, \mathcal{L}^{(i)}\}_{i=1}^{m}$$

Result of the random search stored in $\mathcal{H}_{Random}$ is a collection of all possible hyperparameter configurations and their corresponding loss, which is used to select the best preforming hyperparameter.

**1.5: Selection of best hyperparameter set**: Select the best parameter configuration $\hat{\gamma}_{Random}$, that minimizes the validation loss.

$$\hat{\gamma}_{Random} = arg_{\gamma \in \gamma_{Random}} \min \mathcal{L}(f_\gamma(x_{validation}, y_{validation}))$$

This step focuses on identifying best hyperparameter set $\hat{\gamma}_{Random}$, that achieved minimal loss which is an initial estimate of optimal hyperparameter.

**Step 2. Grid Search**:

**2.1: Define the Grid search range**: After identifying $\hat{\gamma}_{Random}$, define a local grid search range around hyperparameter in $\hat{\gamma}_{Random}$:

$$\Gamma_{Grid} = \{\gamma_j \in [\hat{\gamma}_{random} - \Delta_j , \hat{\gamma}_{random} + \Delta_j ]\}_{j=1}^{n}$$

Where $\Delta_j$ defines the grid refinement range around optimal random hyperparameter $\hat{\gamma}_{random}$.

**2.2: Exhaustive Grid Search**: Perform an exhaustive grid search over the narrower hyperparameter space $\Gamma_{Grid}$, for each set of hyperparameters $\gamma_j$ in the grid:

$$f_{\gamma_j} = Train_ML_Classifier(\mathcal{D}_{Training}, \gamma_j)$$

Here in this step, unlike random search where each configuration are not sampled, grid search exhaust each configuration of $\Gamma_{Grid}$ and evaluates the results.

**2.3: Compute the validation loss of each $f_{\gamma_j}$ grid search model**:

$$\mathcal{L}_{Grid}(\gamma_j) = \mathcal{L}(f_{\gamma_j}(x_{Validation}), y_{Validation})$$

This step quantifies the validation loss results on each $\Gamma_{Grid}$ configurations of hyperparameters.

**2.4: Storing Grid Search Result**:

Stored grid search results in structured form

$$\mathcal{H}_{Grid} = \{\gamma_j, \mathcal{L}_{Grid}(\gamma_j) \mid \gamma_j \in \Gamma_{Grid}\}$$

$\mathcal{H}_{Grid}$ holds the record of all grid search configurations along with their validation losses which is further used to identify best performing hyperparameter set configuration.

**2.5: Selecting the best parameter set**: Selects best parameters $\gamma^*$ which has minimal validation loss.

$$\gamma^* = arg_{\gamma_j \in \mathcal{H}_{Grid}} \min(\mathcal{L}_{Grid}(\gamma_j))$$

In this step, $\gamma^*$ is selected as a best hyperparameter configuration with minimum validation loss among all other configuration in $\mathcal{H}_{Grid}$.

**Step 3. Final Model Training and Evaluation**:

**3.1: Train the final ML Classifier**: Using the optimal hyperparameter $\gamma^*$, train the final ML classifier on entire training set $\mathcal{D}_{Training}$:

$$f_{\gamma^*} = Train_ML_Classifier(\mathcal{D}_{Training}, \gamma^*)$$

Where $f_{\gamma^*}$ is the final ML classifier trained with optimal hyperparameter on entire training set.

**3.2: Test Set Evaluation**: Evaluate the final ML classifier $f_{\gamma^*}$, on test set $\mathcal{D}_{Testing}$, and compute the performance of test data:

$$\mathcal{L}_{Test} = \mathcal{L}(f_{\gamma^*}(X_{Test}, Y_{Test}))$$

Where $\mathcal{L}_{Test}$ represents minimal validation loss using optimal hyperparameter $\gamma^*$.

## 4. Experimental Results and Discussion

This section evaluates the proposed Randomized-Grid Search hyperparameter tuning approach against traditional Grid Search and Random Search for CVDs classification. We employ a variety of machine learning classifiers, such as Logistic Regression (LR), Naïve Bayes (NB), K-Nearest Neighbour (KNN), Support Vector Machine (SVM), Neural Network (NN), and Decision Tree (DT). In assessing performance, we evaluate the classification accuracy (%) and the ROC-AUC metric along with the time taken for training models in different settings.

### 4.1 Dataset Description

The dataset used in this research study is sourced from the UCI Heart Disease repository, which have been commonly used for research focusing on predicting CVDs [31]. It holds the data gathered from a variety of medical examinations and clinical assessments. The dataset presented in Table 2, includes both, numerical and categorical features, such as demographic details like age and sex along with the clinical records like cholesterol level, blood pressure, and heart rate. Many other features like ECG, thallium stress tests are also included in this dataset to capture broader aspects of an individual. Also, the dataset provides the label for the indication whether patients have the heart disease or not. This dataset is widely used in the development of AI models, making it as a benchmark dataset for studying the risk factors associated with the CVDs.

| **Number** | **Variable Name** | **Description/Interpretation** | **Range** |
|---|---|---|---|
| 1 | Age | Patient's age (Years) | 29 to 77 |
| 2 | Sex | Gender of patient | 0= Female<br>1= Male |
| 3 | Chest pain type | Type of chest pain experienced in patients | 0= TA (Typical Angina)<br>1= ATA (Atypical Angina)<br>2= NAP (Non Anginal Pain)<br>3= ASY(Asymptotic) |
| 4 | Resting blood pressure | Resting blood pressure mmHg | 94 to 200 |
| 5 | Total cholesterol | Cholesterol level in mg/dl | 126-564(in mg/dl) |
| 6 | Fasting blood sugar (glucose level) | Whether fasting blood sugar >120mg/dl | 0= False<br>1= True |
| 7 | ECG(Electrocardiography) at rest | Resting ECG results, which indicate electrical activity of the heart and can indicate abnormal pattern | 0= Normal<br>1= ST wave abnormality<br>2= Left Ventricular Hypertrophy |
| 8 | Maximum heart rate achieved | Maximum heart rate achieved during the exercise | 71 to 202 bpm |
| 9 | Exercising angina | Demonstrates if a patients have exercise induced angina | 0= No<br>1= Yes |
| 10 | Old peak | ST depression induced by exercise relative to rest | 0 to 6.2 |
| 11 | slope | The slope of the peak exercise ST segment | 0= Flat<br>1= Up<br>2= Down |
| 12 | CA (Number of major vessels) | Number of major vessels (0-3) coloured by fluoroscopy | 0 to 3 |
| 13 | Thal (Thallium) | Thallium Stress Test | 0 = Null<br>1= Normal |

| | | | 2= Fixed Defect<br>3= Reversible Affect |
|---|---|---|---|
| 14 | Heart Disease | Binary Class | 0= No<br>1= Yes |

***Table 2****: Detailed characteristics of UCI Heart Disease Dataset used in this study.*

### 4.2 Performance Evaluation Metrics:

To evaluate the performance of ML models, performance evaluation metrics such as accuracy, precision, recall, F1 score and ROC-AUC are used which are further defined. For basic notations of confusion matrix cases are explained in Table 3.

| | |
|---|---|
| TP (True Positive) | The number of cases where model correctly identifies the presence of Heart Disease |
| TN (True Negative) | The number of cases where model correctly identifies the absence of Heart Disease |
| FP (False Positive) | The number of cases where model mistakenly identifies as presence of Heart Disease but actually it is absence |
| FN (False Negative) | The number of cases where model mistakenly identifies as the absence of Heart Disease but it is present actually |
| TPR (True Positive Rate) | Higher value of TPR shows that the model is effective at identifying the presence of Heart Disease |
| FPR (False Positive Rate) | Model with Lower FPR is required to say that very few cases of without Heart Disease identifies as presence of Heart Disease |

***Table 3****: Interpretation of Terms used in performance metrics.*

Different performance evaluation metrics for classification model are defined along with the mathematical formula:

A. Accuracy: The ratio of correctly predicted observations and the total observations.

$$Accuracy = \frac{TP + TN}{\mathrm{TP + TN + FP + FN}}$$

B. Precision: Ratio of correctly predicted positive observations to the total predicted positive observation

$$Precision = \frac{TP}{TP + FP}$$

C. Recall: It is the ratio of correctly predicted positive observation to all observation in actual class

$$Recall = \frac{TP}{TP + FN}$$

D. F1 score: It is the weighted average of precision and recall

$$\text{F1 Score} = 2\left(\frac{Precision * Recall}{Precision + Recall}\right)$$

E. ROC (Receiver Operating Characteristics): A visual graph showing the performance model at all classification thresholds plotting true positive rates against false positive rates.

$$\text{TPR(True Positive Rate)} = \frac{TP}{TP + FN}$$

$$\text{FPR(False Positive Rate)} = \frac{FP}{FP + TN}$$

F. AUC (Area Under the Curve): The models ability to classify between positive and negative class

$$\text{AUC} = \sum_{i=1}^{n-1} \left(\frac{TPR_{i+1} + TPR_i}{2}\right)(FPR_i - FPR_{i+1})$$

### 4.3 Hyperparameters Used in this Study

Since the proposed Randomized-Grid Search framework relies on evaluating models over a controlled and well-defined search space, the hyperparameter ranges for all selected classifiers were carefully specified prior to experimentation. The chosen ranges include widely accepted values while ensuring adequate exploration of both low- and high-complexity configurations. The set of tuned hyperparameters for each model, along with their explored ranges, is reported in Table 4.

| Model | Hyperparameter | Range |
|---|---|---|
| **Logistic Regression** | Penalty | l2, l1, elasticnet, none |
| | C (log scale) | 0.0001 - 1000 |
| | solver | lbfgs, liblinear, saga, newton-cg |
| | Class weight | None, balanced |
| **Naive Bayes** | Variable Smoothing | **0.000000000001 - 0.000001** |
| **K-Nearest Neighbour** | N-neighbours | 1 - 200 |
| | weights | uniform, distance |
| | metric | minkowski, cosine |
| | p | 1, 2, 3, 4, 5 |
| **Support Vector Machine** | c | 0.0001 - 1000 |
| | class weight | None, balanced |
| | kernel | linear, rbf, poly, sigmoid |
| **Neural Network** | hidden layer sizes | number of layers: 1 - 3; units per layer: 16 - 512 |
| | activation functions | relu, tanh |
| | alpha | 0.000001 - 0.01 |
| | learning rates | constant, invscaling, adaptive |
| | batch size | 32 - 512 |

| | early stopping | True, False |
|---|---|---|
| **Decision Tree** | Criterion | gini, entropy, log_loss |
| | Splitter | best, random |
| | Maximum Depth | 1 – 50, or None |
| | Minimum Sample Leaf | 2 – 20 |
| | Minimum Sample per Leaf | 1 – 20 |
| | Minimum Weight Fraction per Leaf | 0.0 – 0.5 |
| | Maximum Features | None, sqrt, log2, 0.1 – 1.0 |
| | Maximum Leaf Nodes | 10 – 200, or None |
| | Minimum Purity Decrease | 0.0 – 0.1 |
| | Cost Complexity Pruning Alpha | 0.00001 – 0.1 |
| **Random Forest** | n_estimators | 50 - 1000 |
| | max_depth | 1 - 50, or None |
| | min_samples_split | 2 - 20 |
| | min_samples_leaf | 1 - 20 |
| | max_features | sqrt, log2, 0.3 - 1.0 |
| | bootstrap | True, False |
| | class_weight | None, balanced |
| | criterion | gini, entropy, log_loss |
| | oob_score | True, False |
| | max_samples | 0.5 - 1.0 |
| **XGBoost** | n_estimators | 50 – 1000 |
| | max_depth | 2 – 15 |
| | learning_rate | 0.01 – 0.3 |
| | min_child_weight | 1 – 10 |
| | subsample | 0.5 – 1.0 |
| | colsample_bytree | 0.5 – 1.0 |
| | gamma (min_split_loss) | 0 – 5 |
| | reg_lambda (L2 regularization) | 0 – 10 |
| | reg_alpha (L1 regularization) | 0 – 5 |
| | scale_pos_weight | 1 – 10 |

***Table 4:*** *Detailed Hyperparameter Configuration (Search Space and Ranges) Used for Tuning Classical ML Models and Advanced Ensemble Methods.*

### 4.4 Results and Discussion

The result section is presented in terms of performance along with the training time variability analysis taken by each models. Along with the performance comparison and benefits of proposed models, incremental analysis and discussion is also included. As presented in Table 5, a comparative classification performance over all models.

The proposed Randomized-Grid Search method consistently provides very competitive results for most of the classifiers. This method nicely combines global search using random sampling and local search using fine grid search in the good regions. This technique is increases the ability to identify stable hyperparameter settings without the high computational overhead of brute force search.

| Model | Grid Search Accuracy (ROC-AUC) | Random Search Accuracy (ROC-AUC) | Randomized-Grid Search Accuracy (ROC-AUC) |
|---|---|---|---|
| Logistic Regression | 80.33 (0.8636) | 79.89 (0.8578) | 83.02 (0.8755) |
| Naive Bayes | 81.97 (0.8755) | 81.78 (0.8682) | 81.93 (0.8760) |
| K-Nearest Neighbour | 78.69 (0.8853) | 75.41 (0.8810) | 85.08 (0.8890) |
| Support Vector Machine | 83.44 (0.8571) | 83.04 (0.8522) | 87.96 (0.8712) |
| Neural Network | 83.03 (0.8734) | 82.67 (0.8658) | 83.49 (0.8734) |
| Decision Tree | 77.28 (0.7029) | 77.25 (0.7029) | 81.43 (0.8074) |
| Random Forest | 83.44 (0.9048) | 82.67 (0.9134) | 83.49 (0.9134) |
| XGBoost | 81.40 (0.8929) | 83.79 (0.8961) | 84.30 (0.8994) |

***Table 5:*** *Comparative performance over proposed randomized grid search over grid search and random search.*

The logistic regression model achieved 83.02% accuracy using randomised grid search. The corresponding ROC-AUC of the proposed randomised grid search is 0.8755. This method outperformed grid search (accuracy: 80.33%, ROC-AUC: 0.8636) and random search (accuracy: 79.89%, ROC-AUC: 0.8578). This indicates that the proposed approach can identify better hyperparameters (such as regularisation strength and solver settings) even for simple linear models, which leads to better generalisation. Results for Naïve Bayes (NB) were consistent across different tuning methods. Randomized-Grid Search achieved 81.93% (0.8760) which is almost similar to Grid Search with 81.97% (0.8755) and slightly better than Random Search with 81.78% (0.8682). This is expected because of the relatively low number of parameters that can be tuned in NB. The performance of NB mainly depends on assumptions about the data distribution rather than tuning many hyperparameters.

For KNN, Randomised Grid Search has an accuracy of 85.08% and a ROC-AUC of 0.8890 compared to Grid Search with an accuracy of 78.69% and a ROC-AUC of 0.8853 and Random Search with an accuracy of 75.41% and a ROC-AUC of 0.8810 which is an observable improvement. This means that KNN is very sensitive to careful hyperparameter tuning, like the number of neighbours (k), the distance metric, and the weighting scheme. The proposed method enables to explore selectively within favourable hyperparameter regions which is critical in neighbourhood based learning as performance can vary widely with small parameter changes.

Similarly, SVM method has shown its best performance by using Randomized-Grid Search with an accuracy of 87.96% and ROC-AUC of 0.8712. This performance exceeds the Grid Search results, which provides accuracy of 83.44% and ROC-AUC of 0.8571, and Random Search, which provides accuracy of 83.04% and ROC-AUC of 0.8522. The kernel and the regularisation parameters (C and $\gamma$) configurations are deeply influential in the effectiveness of SVM. The suggested method balances effectively between random exploration and systematic grid refinement for obtaining optimal hyperparameter configurations. The observed increase of the Neural Network performance was relatively modest but consistent with Randomized-Grid Search (accuracy 83.49%, ROC-AUC 0.8734), Grid Search (accuracy 83.03%, ROC-AUC 0.8734) and Random Search (accuracy 82.67%,

ROC-AUC 0.8658). Many neural networks perform well on small datasets and across various configurations, and the proposed method still yields higher accuracy than the baseline random tuning.

The most significant improvement was achieved by the Decision Tree where Randomized-Grid Search reached an accuracy of 81.43% and ROC-AUC of 0.8074. On the other hand, Grid Search and Random Search gave almost same results with 77.28% accuracy and 0.7029 ROC-AUC and 77.25% accuracy and 0.7029 ROC-AUC respectively. The results suggest that the performance of decision trees is significantly affected by hyperparameters controlling the complexity of trees and the constraints on splits, whereas the proposed tuning method greatly enhances both accuracy and discrimination power. The increase in ROC-AUC for Decision Tree indicates improved class separability on the evaluated dataset, suggesting that the proposed tuning strategy is particularly useful for models whose performance strongly depends on split-related hyperparameters.

The proposed Randomized-Grid Search also showed favourable performance for ensemble-based classifiers. For Random Forest, the method achieved the highest accuracy of 83.49% and a ROC-AUC of 0.9134. This indicates an improvement in accuracy compared with both Grid Search and Random Search, while the ROC-AUC improved over Grid Search and remained equal to Random Search. For XGBoost, Randomized-Grid Search achieved the highest accuracy of 84.30% and ROC-AUC of 0.8994, outperforming both Grid Search and Random Search. The findings indicate that the suggested tuning method is effective for both conventional classifiers and advanced ensemble learners. The enhancement of discrimination performance leads to increased accuracy in predictions.

| Model | Grid Search (Seconds) | Random Search (Seconds) | Randomized-Grid Search (Second) |
|---|---|---|---|
| Logistic Regression | 2.8187 | 0.0384 | 0.8985 |
| Naive Bayes | 0.8240 | 0.0930 | 0.0535 |
| K-Nearest Neighbour | 5.0 | 0.2932 | 2.6115 |
| Support Vector Machine | 10.5500 | 6.7820 | 6.1294 |
| Neural Network | 91.7900 | 20.8000 | 70.9886 |
| Decision Tree | 14.5200 | 2.5400 | 7.1720 |
| Random Forest | 42.00 | 11.0637 | 17.7872 |
| XGBoost | 599.8878 | 18.2559 | 13.3861 |

***Table 6:*** *Comparison of training time taken using grid search, random search and proposed randomized-grid search.*

The evaluation of the proposed framework included an assessment of computational efficiency, alongside predictive performance, as detailed in Table 6. In general, the approach with the exhaustive evaluation of the entire hyperparameter space required the most time, particularly for computationally intensive models. Random Search was the fastest in most of the models, but it may miss optimal hyperparameter regions due to its lack of systematic refinement. The proposed Randomized-Grid Search is a good compromise: it significantly reduces computation costs with respect to Grid Search, while providing performance which is consistently better than Random Search and often better than Grid Search.

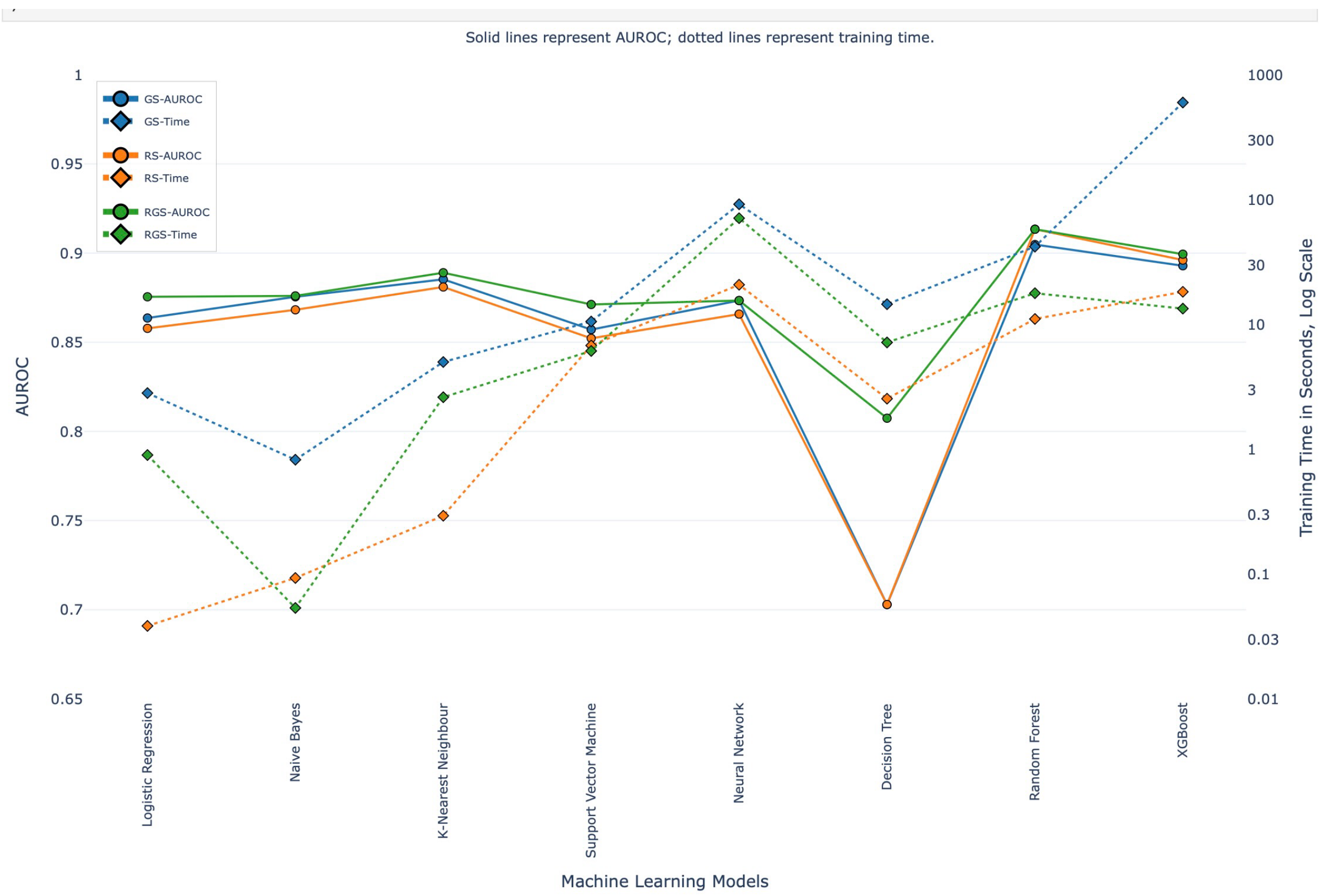

***Fig. 5:*** *Dual-axis comparison of AUROC and training time across Grid Search, Random Search, and Randomized-Grid Search. Solid lines represent AUROC values on the left y-axis, whereas dotted lines represent training time in seconds on the logarithmic right y-axis.*

The dual-axis visualization in Fig. 5 provides a combined assessment of predictive discrimination and computational cost across all evaluated models. The solid lines indicate that Randomized-Grid Search generally achieved improved or competitive AUROC values, with noticeable gains for Logistic Regression, KNN, SVM, Decision Tree, and XGBoost. For Random Forest, Randomized-Grid Search maintained the same AUROC as Random Search while improving over Grid Search. The dotted lines show training time on a logarithmic scale, which enables comparison across models with substantially different runtimes. The visualization further shows that Randomized-Grid Search consistently reduced runtime compared with Grid Search, especially for computationally expensive models such as Neural Network, Random Forest, and XGBoost. However, compared with Random Search, the runtime pattern was model-dependent because the proposed method includes an additional local grid-refinement phase. Overall, Fig. 5 supports the observation that Randomized-Grid Search offers a practical trade-off between AUROC performance and computational efficiency.

For instance, in the context of Decision Tree, Grid Search took 14.52 seconds, Random Search took only 2.54 seconds, and Randomized-Grid Search took 7.17 seconds. This shows a significant time reduction as compared to Grid Search with achieving the highest accuracy and ROC-AUC. The Grid Search method for SVM took 10.55 seconds while the Randomized-Grid Search method cut the runtime substantially to 6.13 seconds and improved the accuracy to 87.96%. When comparing Grid Search and Randomized-Grid Search in the scope of Neural Networks, it showed a significant decrease in the tuning effort, reducing the time from 91.79 seconds to 70.99 seconds while improving the prediction performance. Similarly, though the Grid Search method took 5.0 seconds, the Randomized-Grid Search gave significantly better accuracy in 2.61 seconds. The comparison of computation time for ensemble classifiers also shows the efficiency of the proposed Randomized-Grid Search (RGS). The Grid Search method took 42.00 seconds to complete while both Random Search and RGS significantly decreased the run-time to 11.0637 seconds and 17.7872 seconds respectively. This shows a notable improvement over traditional grid based tuning methods. XGBoost with Grid Search The computational cost was significantly high taking 599.8878 seconds. Random Search, on the other hand, finished in 18.2559 seconds and RGS was the most efficient method requiring only 13.3861 seconds. Interestingly, RGS exhibited the shortest runtime among the other two strategies for XGBoost, and also improved the classification performance, confirming the effectiveness of RGS as a method for optimizing resource-consuming ensemble models.

| Model | Accuracy margin (Randomized-Grid Search Vs Grid Search) | ROC-AUC margin (Randomized-Grid Search Vs Grid Search) | Accuracy margin (Randomized-Grid Search Vs Random Search) | ROC-AUC margin (Randomized-Grid Search Vs Random Search) |
| --- | --- | --- | --- | --- |
| Logistic Regression | +2.69 | +0.0119 | +3.13 | +0.0177 |
| Naive Bayes | −0.04 | +0.0005 | +0.15 | +0.0078 |
| K-Nearest Neighbour | +6.39 | +0.0037 | +9.67 | +0.0080 |
| Support Vector Machine | +4.52 | +0.0141 | +4.92 | +0.0190 |
| Neural Network | +0.46 | +0.0000 | +0.82 | +0.0076 |
| Decision Tree | +4.15 | +0.1045 | +4.18 | +0.1045 |
| Random Forest | +0.05 | +0.0086 | +0.82 | +0.0000 |
| XGBoost | +2.9 | +0.0065 | +0.51 | +0.0033 |

***Table** 7: Variation in performance of Randomized-Grid Search and other two searches: Random Search and Grid Search with given accuracy and ROC-AUC.*

The performance margins of the proposed Randomized-Grid Search in comparison to Grid Search and Random Search are detailed in Table 7, utilizing accuracy and ROC-AUC as metrics. In comparison to Grid Search, Randomized-Grid Search enhances accuracy for Logistic Regression by 2.69%, for K-Nearest Neighbour by 6.39%, for Support Vector Machine by 4.52%, for Neural Network by 0.46%, and for Decision Tree by 4.15%. A minimal decrease of 0.04% is noted for Naïve Bayes. The margins for ROC-AUC compared to Grid Search are generally favorable across the majority of models.

Specifically, there are enhancements of 0.0119 for Logistic Regression, 0.0005 for Naïve Bayes, 0.0037 for K-Nearest Neighbour, 0.0141 for Support Vector Machine, and 0.1045 for Decision Tree, whereas the Neural Network exhibits no variation.

In comparison to Random Search, Randomized-Grid Search enhances accuracy for all classifiers assessed,

achieving improvements of 3.13% for Logistic Regression, 0.15% for Naïve Bayes, 9.67% for K-Nearest Neighbour, 4.92% for Support Vector Machine, 0.82% for Neural Network, and 4.18% for Decision Tree. The variable behaviour of ROC-AUC are consistently positive changing from 0.0076 to 0.0190 across the majority of different models with a peak of 0.1045 observed for the Decision Tree model. Conclusively, Decision Tree demonstrates the most significant ROC-AUC difference margin while compared to both baselines: highlighting a notable enhancement with the suggested tuning approach. For ensemble classifiers, Randomized-Grid Search also showed favourable performance margins. In Random Forest, the proposed method improved accuracy by 0.05% compared with Grid Search and by 0.82% compared with Random Search, while improving ROC-AUC over Grid Search by 0.0086 and maintaining the same ROC-AUC as Random Search. For XGBoost, Randomized-Grid Search improved accuracy by 2.90% and ROC-AUC by 0.0065 compared with Grid Search, and improved accuracy by 0.51% and ROC-AUC by 0.0033 compared with Random Search.

| Model | Time taken by Grid Search | Time taken by Random Search | Time taken by Randomized-Grid Search | Time difference between (Randomized-Grid Search and Grid Search) | % Time saving vs Grid Search | Time difference between (Randomized-Grid Search and Random Search) |
|---|---|---|---|---|---|---|
| Logistics Regression | 2.8187 | 0.0384 | 0.8985 | -1.9202 | 68.1% Faster | +0.8601 |
| Naive Bayes | 0.8240 | 0.0930 | 0.0535 | -0.7705 | 93.5% Faster | -0.0395 |
| KNN | 5.0000 | 0.2932 | 2.6115 | -2.3885 | 47.8% Faster | +2.3183 |
| SVM | 10.5500 | 6.7820 | 6.1294 | -4.4206 | 41.9% Faster | -0.6526 |
| Neural Network | 91.7900 | 20.8000 | 70.9886 | -20.8014 | 22.7% Faster | +50.1886 |
| Decision Tree | 14.5200 | 2.5400 | 7.1720 | -7.3480 | 50.6% Faster | +4.6320 |
| Random Forest | 42.00 | 11.0637 | 17.7872 | -24.2128 | 57.6% Faster | +6.7235 |
| XGBoost | 599.8878 | 18.2559 | 13.3861 | -586.5017 | 97.8% Faster | -4.8698 |

***Table 8****: Comparison of computation time of Randomized-Grid Search, Grid Search and Random Search*

The computation-time comparison among Grid Search, Random Search, and the proposed Randomized-Grid Search is illustrated in Table 8. In comparison to Grid Search, Randomized-Grid Search significantly decreases the runtime for Naïve Bayes from 0.8240 seconds to 0.0535 seconds, resulting in a reduction of 0.7705 seconds or 93.5%. The runtime for KNN has been reduced from 5.0000 seconds to 2.6115 seconds, resulting in a decrease of 2.3885 seconds, which corresponds to a 47.8% reduction. The runtime for SVM has been reduced from 10.5500 seconds to 6.1294 seconds, indicating a decrease of 4.4206 seconds, which corresponds to a 41.9% improvement. The runtime for the Neural Network has decreased from 91.7900 seconds to 70.9886 seconds, indicating a reduction of 20.8014 seconds or 22.7%. The runtime for the Decision Tree has been reduced from 14.5200 seconds to 7.1720 seconds, reflecting a decrease of 7.3480 seconds, which is a 50.6% reduction. Compared to the Random Search, Randomized-Grid Search has time efficiency for Naïve Bayes, which is reduced by 0.0395 seconds, and for SVM, which is reduced by 0.6526 seconds. Meanwhile, the computation time is increased by 2.3183 s, 50.1886 s, and 4.6320 s for KNN, Neural Network, and Decision Tree, respectively. This increase can be linked with the refinement phase introduced after the random exploration in the proposed strategy.

## 5. Conclusion

We propose a strategy of hyperparameter optimization that combines random exploration with a targeted local grid refinement. The analysis of UCI Heart Disease dataset showed robust and reliable classification results over various machine learning classifiers. The results showed a significant improvement in the Decision Tree performance with 81.43% accuracy and 0.8074 ROC-AUC compared to 77.28% and 0.7029 previously obtained by Grid Search. The identical tuning method yielded favourable results for models with greater capacity, such as Random Forest achieving a ROC-AUC of 0.9134 and XGBoost reaching an accuracy of 84.30% along with a ROC-AUC of 0.8994. In addition to enhancing predictive performance, the approach significantly lowered the computational demands compared to exhaustive Grid Search for the majority of models. The improvement in efficiency is particularly striking for XGBoost, as the tuning duration decreased from 599.8878 seconds using Grid Search to just 13.3861 seconds with Randomized-Grid Search, all while achieving enhanced performance. The suggested Randomized-Grid Search presents a viable tuning option for healthcare machine learning pipelines, achieving an improved equilibrium between accuracy-focused optimization and time-efficient exploration compared to traditional Grid Search and Random Search methods.

## Reference

1. 'Cardiovascular diseases (CVDs)'. Accessed: Nov. 18, 2024. [Online]. Available: https://www.who.int/news-room/fact-sheets/detail/cardiovascular-diseases-(cvds)
2. G. A. Roth *et al.*, 'Global Burden of Cardiovascular Diseases and Risk Factors, 1990–2019: Update From the GBD 2019 Study', *Journal of the American College of Cardiology*, vol. 76, no. 25, pp. 2982–3021, Dec. 2020, doi: 10.1016/j.jacc.2020.11.010.
3. S. Darrab, D. Broneske, and G. Saake, 'Exploring the predictive factors of heart disease using rare association rule mining', *Sci Rep*, vol. 14, no. 1, p. 18178, Aug. 2024, doi: 10.1038/s41598-024-69071-6.
4. Y. Zheng and X. Hu, 'Healthcare predictive analytics for disease progression: a longitudinal data fusion approach', *J Intell Inf Syst*, vol. 55, no. 2, pp. 351–369, Oct. 2020, doi: 10.1007/s10844-020-00606-9.
5. T. Nagamine, B. Gillette, J. Kahoun, R. Burghaus, J. Lippert, and M. Saxena, 'Data-driven identification of heart failure disease states and progression pathways using electronic health records', *Sci Rep*, vol. 12, no. 1, p. 17871, Oct. 2022, doi: 10.1038/s41598-022-22398-4.
6. Y. Zhang, J. R. Golbus, E. Wittrup, K. D. Aaronson, and K. Najarian, 'Enhancing heart failure treatment decisions: interpretable machine learning models for advanced therapy eligibility prediction using EHR data', *BMC Medical Informatics and Decision Making*, vol. 24, no. 1, p. 53, Feb. 2024, doi: 10.1186/s12911-024-02453-y.
7. A. Esteva *et al.*, 'A guide to deep learning in healthcare', *Nat Med*, vol. 25, no. 1, pp. 24–29, Jan. 2019, doi: 10.1038/s41591-018-0316-z.
8. E.J. Topol, 'High-performance medicine: the convergence of human and artificial intelligence', *Nat Med*, vol. 25, no. 1, pp. 44–56, Jan. 2019, doi: 10.1038/s41591-018-0300-7.
9. C. Martins *et al.*, 'Identifying subgroups in heart failure patients with multimorbidity by clustering and network analysis', *BMC Medical Informatics and Decision Making*, vol. 24, no. 1, p. 95, Apr. 2024, doi: 10.1186/s12911-024-02497-0.
10. K. W. Johnson *et al.*, 'Artificial Intelligence in Cardiology', *J Am Coll Cardiol*, vol. 71, no. 23, pp. 2668–2679, Jun. 2018, doi: 10.1016/j.jacc.2018.03.521.
11. H. Sadr, A. Salari, M. T. Ashoobi, and M. Nazari, 'Cardiovascular disease diagnosis: a holistic approach using the integration of machine learning and deep learning models', *European Journal of Medical Research*, vol. 29, no. 1, p. 455, Sep. 2024, doi: 10.1186/s40001-024-02044-7.
12. X. Sun, Y. Yin, Q. Yang, and T. Huo, 'Artificial intelligence in cardiovascular diseases: diagnostic and therapeutic perspectives', *European Journal of Medical Research*, vol. 28, no. 1, p. 242, Jul. 2023, doi: 10.1186/s40001-023-01065-y.

13. M. A. Naser, A. A. Majeed, M. Alsabah, T. R. Al-Shaikhli, and K. M. Kaky, 'A Review of Machine Learning's Role in Cardiovascular Disease Prediction: Recent Advances and Future Challenges', *Algorithms*, vol. 17, no. 2, Art. no. 2, Feb. 2024, doi: 10.3390/a17020078.
14. P. Mathur, S. Srivastava, X. Xu, and J. L. Mehta, 'Artificial Intelligence, Machine Learning, and Cardiovascular Disease', *Clin Med Insights Cardiol*, vol. 14, p. 1179546820927404, Jan. 2020, doi: 10.1177/1179546820927404.
15. A. Tiwari, A. Chugh, and A. Sharma, "Ensemble framework for cardiovascular disease prediction," Computers in Biology and Medicine, vol. 146, p. 105624, 2022, doi: 10.1016/j.compbiomed.2022.105624
16. T. Vivekanandan and S. J. Narayanan, "A Hybrid Risk Assessment Model for Cardiovascular Disease Using Cox Regression Analysis and a 2-means Clustering Algorithm," Computers in Biology and Medicine, vol. 116, p. 103400, 2019, doi: 10.1016/j.compbiomed.2019.103400
17. Z. I. Attia *et al.*, 'An artificial intelligence-enabled ECG algorithm for the identification of patients with atrial fibrillation during sinus rhythm: a retrospective analysis of outcome prediction', *The Lancet*, vol. 394, no. 10201, pp. 861–867, Sep. 2019, doi: 10.1016/S0140-6736(19)31721-0.
18. M. Fernández-Delgado, E. Cernadas, S. Barro, and D. Amorim, 'Do we Need Hundreds of Classifiers to Solve Real World Classification Problems?', *Journal of Machine Learning Research*, vol. 15, no. 90, pp. 3133–3181, 2014.
19. P. Probst, M. N. Wright, and A.-L. Boulesteix, 'Hyperparameters and tuning strategies for random forest', *WIREs Data Mining and Knowledge Discovery*, vol. 9, no. 3, p. e1301, 2019, doi: 10.1002/widm.1301.
20. S. Singhal and M. Kumar, 'A Systematic Review on Artificial Intelligence-Based Techniques for Diagnosis of Cardiovascular Arrhythmia Diseases: Challenges and Opportunities', *Arch Computat Methods Eng*, vol. 30, no. 2, pp. 865–888, Mar. 2023, doi: 10.1007/s11831-022-09823-7.
21. Md. I. Hossain *et al.*, 'Heart disease prediction using distinct artificial intelligence techniques: performance analysis and comparison', *Iran J Comput Sci*, vol. 6, no. 4, pp. 397–417, Dec. 2023, doi: 10.1007/s42044-023-00148-7.
22. J. S. Suri, M. Bhagawati, S. Paul, A. Protogeron, P. P. Sfikakis, G. D. Kitas, N. N. Khanna, Z. Ruzsa, A. M. Sharma, S. Saxena, G. Faa, K. I. Paraskevas, J. R. Laird, A. M. Johri, L. Saba, and M. S. Kalra, "Understanding the bias in machine learning systems for cardiovascular disease risk assessment: The first of its kind review," Computers in Biology and Medicine, vol. 140, p. 105204, 2022, doi: 10.1016/j.compbiomed.2021.105204 .
23. K. Sumwiza, C. Twizere, G. Rushingabigwi, P. Bakunzibake, and P. Bamurigire, 'Enhanced cardiovascular disease prediction model using random forest algorithm', *Informatics in Medicine Unlocked*, vol. 41, p. 101316, Jan. 2023, doi: 10.1016/j.imu.2023.101316.
24. P. Probst, M. Wright, and A.-L. Boulesteix, 'Hyperparameters and Tuning Strategies for Random Forest', Feb. 26, 2019, *arXiv*: arXiv:1804.03515. doi: 10.48550/arXiv.1804.03515.
25. F. Hutter, L. Kotthoff, and J. Vanschoren, Eds., *Automated Machine Learning: Methods, Systems, Challenges*. in The Springer Series on Challenges in Machine Learning. Cham: Springer International Publishing, 2019. doi: 10.1007/978-3-030-05318-5.
26. P. Liashchynskyi and P. Liashchynskyi, 'Grid Search, Random Search, Genetic Algorithm: A Big Comparison for NAS', Dec. 12, 2019, *arXiv*: arXiv:1912.06059. doi: 10.48550/arXiv.1912.06059.
27. J. Snoek, H. Larochelle, and R. P. Adams, 'Practical Bayesian Optimization of Machine Learning Algorithms', Aug. 29, 2012, *arXiv*: arXiv:1206.2944. doi: 10.48550/arXiv.1206.2944.
28. J. Howard and S. Gugger, 'fastai: A Layered API for Deep Learning', Feb. 16, 2020, *arXiv*: arXiv:2002.04688. doi: 10.48550/arXiv.2002.04688.
29. A. H. Victoria and G. Maragatham, 'Automatic tuning of hyperparameters using Bayesian optimization', *Evolving Systems*, vol. 12, no. 1, pp. 217–223, Mar. 2021, doi: 10.1007/s12530-020-09345-2.
30. R. Gomes Mantovani *et al.*, 'Better trees: an empirical study on hyperparameter tuning of classification decision tree induction algorithms', *Data Min Knowl Disc*, vol. 38, no. 3, pp. 1364–1416, May 2024, doi: 10.1007/s10618-024-01002-5.
31. W. S. Andras Janosi, 'Heart Disease'. UCI Machine Learning Repository, 1989. doi: 10.24432/C52P4X.
32. A. K. M. R. Bashar, M. Goudarzi, and C. P. Tsokos, 'A Machine Learning Classification Model for Detecting Prediabetes', *JDAIP*, vol. 12, no. 03, pp. 462–478, 2024, doi: 10.4236/jdaip.2024.123024.
33. M. Chaubey, A. K. Pathak, Marisha, and M. Gupta, 'A Comparative Performance Analysis of Activation Functions for Cardiovascular Disease Detection Using ECG Images', *Advanced Theory and Simulations*, Nov. 2025, doi: 10.1002/adts.202501567.
34. A. K. Pathak, B. K. Dewangan, and M. Gupta, 'Predicting Cardiovascular Diseases Using Machine Learning', in Computational Intelligence in Industry 4.0 and 5.0 Applications, 2023, doi: 10.1201/9781003359951-9.
35. Wang, X., Li, S., Pun, C.-M., Guo, Y., Xu, F., Gao, H., & Lu, H. (2023). A Parkinson's auxiliary diagnosis algorithm based on a hyperparameter optimization method of deep learning. IEEE/ACM Transactions on Computational Biology and Bioinformatics, 21(4), 912–923.
36. Kaur, B. P., Singh, H., Hans, R., Sharma, S. K., Sharma, C., & Hassan, M. M. (2024). A genetic algorithm aided hyper parameter optimization based ensemble model for respiratory disease prediction with explainable AI. *PLOS ONE, 19*(12), e0308015.
37. Lin, C.-M., & Lin, Y.-S. (2024). TPTM-HANN-GA: A novel hyperparameter optimization framework integrating the Taguchi method, an artificial neural network, and a genetic algorithm for the precise prediction of cardiovascular disease risk. Mathematics, 12(9), 1303.
38. G. Dai, Z. Tian, C. Wang, Q. Tang, H. Chen, and Y. Zhang, 'Lightweight Vision Transformer with Lite-AVPSO Hyperparameter Optimization for Agricultural Disease Recognition', *IEEE Internet of Things Journal*, vol. 12, no. 23, pp. 49142–49159, 2025.
39. K. Vanitha, M. T. R., S. Sathea Sree, and S. Guluwadi, 'Deep learning ensemble approach with explainable AI for lung and colon cancer classification using advanced hyperparameter tuning', *BMC Medical Informatics and Decision Making*, vol. 24, no. 1, p. 222, 2024.

**'Ethics, Consent to Participate, and Consent to Publish declarations: not applicable.**

**Author Contribution**

Research concept and design: AP and MG; Collection and/or assembly of data: AP and MC; Data analysis and interpretation: AP; Mathematical Formulation: AP, MC; Data Visualization: AP, MC; Writing the article: AP, MC; Critical revision of the article: AP, MG.

**Data Declaration**

The datasets used during the present study are available publicly from the UCI Machine Learning Repository. The Link of the repository is given as **"https://archive.ics.uci.edu/dataset/45/heart+disease" .**

**Funding**

The author(s) declare that no financial support was received for the research, authorship, and/or publication of this article.

**Conflict of interest**

The authors have no conflicts of interest to declare.